\documentclass[a4paper,fleqn]{cas-sc}

\usepackage[numbers]{natbib}
\usepackage{amsmath}
\usepackage{amssymb}
\usepackage{booktabs}
\usepackage{caption}
\usepackage{graphicx}
\usepackage{hyperref}
\usepackage{multirow}
\usepackage{url}

\begin{document}

\let\WriteBookmarks\relax
\def\floatpagepagefraction{0.7}
\def\textpagefraction{.01}

\shorttitle{Where Does Neural Advantage Arise in Continuous-Time Dynamic Graph Prediction?}
\shortauthors{Yu and Ha}

\title[mode=title]{Where Does Neural Advantage Arise in Continuous-Time Dynamic Graph Prediction?}

\author[1]{Minwoo Yu}
\ead{snowypainter@konkuk.ac.kr}

\author[1]{Young-guk Ha}
\cormark[1]
\ead{ygha@konkuk.ac.kr}

\cortext[cor1]{Corresponding author}

\affiliation[1]{organization={Smart Computing Laboratory, Department of Computer Science \& Engineering, Konkuk University},
            addressline={120 Neungdong-ro, Gwangjin-gu},
            city={Seoul},
            postcode={05029},
            country={Republic of Korea}}

\begin{abstract}
Aggregate performance on continuous-time dynamic graphs (CTDGs) combines, in a single score, the portion attributable to known temporal regularities and the additional predictive power of neural models. This study separates the two at the query level. We construct a mechanism-constrained predictor that uses pair recurrence, recency and history position, renewal patterns, and short sequential transitions while learning the compatibility within each mechanism. Across four CTDG datasets, this predictor recovers a substantial portion of the performance of strong neural baselines, and the recovered performance quickly saturates with a small, dataset-specific set of explicit mechanisms.

Neural residuals concentrate on queries for which the positive and negative candidates have similar mechanism-execution profiles. Allowing conditional interactions among mechanisms is more effective than simply reweighting their existing contributions. Conditioning the contribution of one mechanism on the execution state of another recovers 54.9--73.2\% of the original neural-only queries and improves overall paired accuracy on all four datasets. Although the magnitude of the effect varies across datasets, these results show that the performance gap of neural CTDG models need not be treated solely as an opaque difference in representational capacity. At least part of the gap is localized to queries with similar candidate execution profiles and can be functionally explained by conditional coordination among known, low-dimensional mechanisms.
\end{abstract}

\begin{highlights}
\item We separate CTDG performance into known-mechanism coverage and neural residuals.
\item Few explicit temporal mechanisms recover much of neural benchmark performance.
\item Neural residuals cluster when candidate mechanism-execution profiles align.
\item Conditional mechanism interactions recover many neural-only successes.
\item The interaction probe improves accuracy rather than merely shifting errors.
\end{highlights}


\maketitle

\section{Introduction}
\label{sec:introduction}

Continuous-time dynamic graph (CTDG) models compress interaction histories into memory states, temporal neighborhoods, or latent representations of event sequences. Models such as TGN~\cite{rossi2020tgn}, TGAT~\cite{xu2020tgat}, GraphMixer~\cite{cong2023graphmixer}, and DyGFormer~\cite{yu2023dygformer} have developed different approaches to memory updates, temporal encoding, neighborhood aggregation, and sequence modeling. Yet aggregate performance alone cannot determine whether the gains of increasingly complex architectures come from better use of known temporal regularities or from additional predictive structure beyond those regularities.

This issue has been raised from several directions. EdgeBank~\cite{poursafaei2022dgb} showed that simply memorizing past edges can compete with established models, while GraphMixer~\cite{cong2023graphmixer} demonstrated the strength of a simplified architecture. Recent heuristic-based predictors~\cite{dileo2024heuristics,cornell2025heuristics,kondrup2025base3} have likewise achieved strong CTDG performance using signals such as recurrence, recency, popularity, and temporal co-occurrence. In parallel, TGB-Seq~\cite{yi2025tgbseq} and Hayes et al.~\cite{hayes2025learn} weakened recurrent edges or isolated individual graph properties to test which temporal structures existing models actually learn. Together, these findings establish both the predictive strength of simple temporal mechanisms and the need to examine the source of neural performance separately.

One question nevertheless remains: \emph{Where exactly does the additional predictive power of neural CTDG models arise beyond known temporal interaction mechanisms, and what explains it?}

In this study, \emph{known temporal interaction mechanisms} refer to a prespecified set of mechanisms executable over the local interaction history preceding a query: recurrence of the same pair, recency and history position, renewal patterns, and short sequential transitions.

Answering the question requires a reference model that fixes the available mechanisms while learning temporal and candidate compatibility within each mechanism. Such a model must be more flexible than a fixed heuristic yet have a clearer computational scope than a general neural predictor, so that the predictive power of a mechanism can be separated from the rigidity of its implementation and the source of recovered performance can be traced.

We use a simple neuro-symbolic predictor, LiFTER, to establish this known-mechanism boundary. LiFTER represents past interactions as facts that preserve entity identity and timestamps, and produces a score when recurrence, history position, renewal, or a short transition is grounded in actual historical facts. Symbolic execution determines which facts participate in each mechanism, while neural parameters learn the compatibility and contribution magnitude of valid executions. The known-mechanism boundary is therefore a learnable reference under an explicit execution and scoring design; a LiFTER error denotes the residual left by this reference.

LiFTER recovers a substantial portion of the performance of strong neural baselines on four CTDG datasets, and this performance quickly saturates with a small, dataset-specific set of mechanisms. We define a \emph{neural residual} as the additional predictive power that remains after accounting for this reference. To observe it at the query level, we define a \emph{neural-only success} as a candidate pair that LiFTER ranks incorrectly but a neural model ranks correctly. Neural-only success is thus the query-level diagnostic unit used to locate and analyze the neural residual.

Neural-only successes occur for every dataset and neural model considered. They concentrate not simply where few mechanisms execute, but where the positive and negative candidates have similar mechanism-execution profiles and are therefore difficult to distinguish structurally with the explicit mechanisms. Residual traces further show cases in which the same mechanisms execute for both candidates but their contributions are insufficiently separated or combined in an unfavorable direction.

We test this explanation by comparing a linear probe that reweights existing mechanism contributions with an interaction probe that conditions the contribution of one mechanism on the execution state of another. Conditional interactions recover a substantial fraction of the original neural-only queries and improve overall paired accuracy on all four datasets, whereas simple reweighting has no consistent effect. Thus, part of the neural performance gap can be localized to queries on which candidates look similar under known mechanisms and functionally explained by conditional coordination among those mechanisms. The magnitude of this effect and its association with the residual regime vary across datasets, so we interpret it as a functional explanation rather than a universal claim about neural internals.

Our contributions are as follows. First, we introduce a diagnostic decomposition that separates predictive performance covered by known temporal interaction mechanisms from the additional predictive gain left to neural models. We call the latter the neural residual and study it at the query level through neural-only successes under matched candidate pairs.

Second, we show that a small number of explicit temporal mechanisms already accounts for a substantial portion of benchmark performance. We further locate neural-only successes in queries where the positive and negative candidates differ little in mechanism executability, rather than merely in queries with few executable mechanisms.

Third, we show that part of the neural residual is recovered more effectively by conditional interactions among existing mechanisms than by simple reweighting. The interaction probe recovers many original neural-only queries while increasing overall paired accuracy, suggesting that some neural advantage can arise from conditional coordination among known mechanisms as well as from additional temporal patterns.

Finally, these findings point to a concrete direction for CTDG model design. Additional neural capacity may be most useful on queries where available mechanisms overlap or conflict and an additive score cannot cleanly distinguish candidates, rather than being devoted solely to adding more temporal patterns.

\section{Related Work}
\label{sec:related_work}

\subsection{CTDG prediction and architectural simplification}

Representative CTDG models differ in how they represent past interactions. TGN~\cite{rossi2020tgn} uses event-driven entity memories and message passing, while TGAT~\cite{xu2020tgat} applies temporal encoding and attention over temporal neighbors. GraphMixer~\cite{cong2023graphmixer} mixes recent interactions with multilayer perceptrons, and DyGFormer~\cite{yu2023dygformer} processes long histories in patches. CRAFT~\cite{yi2025craft} emphasizes entity identity and comparisons of recent histories conditioned on the candidate destination.

These studies seek stronger or more efficient predictive architectures and explain why particular components support performance. Their results do not directly locate where neural models gain beyond simple temporal interaction mechanisms. Our focus is therefore not another general-purpose architecture, but the queries on which this additional gain appears.

Other work explicitly models structures richer than local mechanisms. CAWN~\cite{wang2021caw} represents temporal motifs through causal anonymous walks, HOT~\cite{besta2024hot} incorporates multi-hop neighborhoods and higher-order subgraphs into a Transformer, and TNCN~\cite{zhang2025tncn} improves pairwise representations with temporal common neighbors. These approaches illustrate structures that models may exploit beyond the known local mechanisms considered here.

\subsection{Recurrence, recency, and benchmark evaluation}

Poursafaei et al.~\cite{poursafaei2022dgb} analyzed the prevalence of repeated edges in real dynamic graphs and proposed EdgeBank, which memorizes observed edges. Its strong performance exposed the ease of random negative sampling and motivated stricter evaluation with historical and inductive negatives. The Temporal Graph Benchmark (TGB) of Huang et al.~\cite{huang2023tgb} extended this effort with large-scale datasets and standardized evaluation.

Dileo and Zignani~\cite{dileo2024heuristics} applied common neighbors, Adamic--Adar, resource allocation, and preferential attachment to TGB and showed that traditional link-prediction heuristics can compete with neural models. Cornell et al.~\cite{cornell2025heuristics} ranked candidates by combinations of global and local recency and popularity. Kondrup~\cite{kondrup2025base3} showed that Base3, which combines EdgeBank, popularity, and temporal co-occurrence, remains strong under historical and inductive negative sampling. The strength of simple signals and their combination is therefore established rather than itself our novelty.

What these studies leave unresolved is where the additional predictive power of neural models occurs. Aggregate comparisons between a heuristic and a neural model do not reveal whether they solve the same queries or which histories characterize neural-only successes. We establish a learnable reference that remains within the permitted mechanisms and analyze the residual neural advantage on matched queries.

Romero et al.~\cite{romero2024evaluation} argued that a single aggregate metric inadequately describes model differences and analyzed performance and errors by negative-sampling scheme, node and edge lifetime, and time interval. Their work motivates regime-level evaluation; ours connects regimes directly to the executability of temporal interaction mechanisms and defines residual neural advantage relative to that boundary.

\subsection{Sequential dynamics and direct tests of model capability}

TGB-Seq~\cite{yi2025tgbseq} observes that existing datasets contain many repeated edges and limited complex sequential dynamics. Performance of models including GraphMixer and DyGFormer drops substantially on datasets that reduce edge recurrence and require generalization to unseen edges and sequential context. TGB-Seq thereby demonstrates failures outside recurrence-heavy benchmarks. Our analysis instead separates, within established benchmarks, the queries covered by known mechanisms from those solved only by neural models.

Hayes et al.~\cite{hayes2025learn} use controlled experiments to test whether existing models learn temporal resolution, directionality, density, persistence, periodicity, recency, homophily, and preferential attachment. Their unit of analysis is the relation between a model and an individual graph property. Our unit is the disagreement between a mechanism-constrained predictor and a neural predictor on the same real query. We first measure the performance covered by known mechanisms and then identify the shared historical structure of neural-only successes. The question is thus not whether a neural model can reproduce a property, but where its additional benchmark predictive power remains and what accounts for it.

\subsection{Temporal-graph explanations and neuro-symbolic execution}

T-GNNExplainer~\cite{xia2023tgnnexplainer} and TempME~\cite{chen2023tempme} identify subsets of events or temporal motifs that preserve a trained CTDG prediction. TGIB~\cite{seo2024tgib} and SIG~\cite{fang2024sig} incorporate sparse explanatory subgraphs into prediction. These methods target important events or subgraphs for a given neural prediction. We use fact-level attribution for a different purpose: to verify that the performance of the constrained predictor is actually produced through the prespecified mechanisms.

Neuro-symbolic methods for temporal knowledge graphs treat historical facts differently. Neural-LP~\cite{yang2017neurallp}, TLogic~\cite{liu2022tlogic}, and TILP~\cite{xiong2024tilp} learn or execute rules that satisfy variable bindings and temporal conditions over relational facts. Grounded executions over concrete entities and timestamps can then serve as evidence for candidate scores. CTDG interactions, however, generally lack a semantic relation vocabulary, which prevents a direct transfer of relational rule chains.

Rather than transferring semantic relation rules, we adopt the computational principle of grounded execution. Exact bindings between past events and query entities, recurrence of the same pair, history position, and temporal order determine executability. This constraint fixes the portion attributable to known mechanisms and makes the remaining neural predictive power observable.

\section{Problem Definition}
\label{sec:problem_definition}

The observations form a time-ordered stream of events, each consisting of a source entity, a destination entity, and an occurrence time:

\begin{equation}
\label{eq:event_stream}
\mathcal E=\{(u_i,v_i,t_i)\}_{i=1}^{N}, \qquad t_i\le t_{i+1}.
\end{equation}

In Eq.~\eqref{eq:event_stream}, $(u_i,v_i,t_i)$ denotes an interaction between source $u_i$ and destination $v_i$ at time $t_i$. Events are provided in ascending temporal order.

The prediction target is a query $q=(X,Y,T_q)$ asking whether $X$ and $Y$ interact at time $T_q$. Only events preceding the query are available:

\begin{equation}
\label{eq:history}
\mathcal H_{<T_q}=\{(u_i,v_i,t_i)\mid t_i<T_q\}.
\end{equation}

Using only the history in Eq.~\eqref{eq:history}, a predictor must assign the true destination $Y^+$ for source $X$ a higher score than a negative candidate $Y^-$.

We use two types of negative candidates. A random negative is likely to include an entity that has never interacted with the query source. A historical negative is sampled from previous destinations of $X$, excluding the current true destination. Because both candidates in the latter setting have appeared in the source history, it tests whether recency, renewal, and sequential transitions distinguish the current destination beyond simple exposure.

Our analysis concerns a prespecified set $\mathcal M$ of temporal interaction mechanisms. Let $G_m$ denote the valid executions of mechanism $m$ for a query and its history, and let $c_m$ be the signed contribution of each execution. The candidate score is

\begin{equation}
\label{eq:mechanism_score}
s(q)=b+\sum_{m\in\mathcal M}\sum_{g\in G_m(q,\mathcal H_{<T_q})}c_m(g,q).
\end{equation}

Equation~\eqref{eq:mechanism_score} defines the analytical boundary: except for the prior $b$, every score component must arise from an actual execution of a mechanism in $\mathcal M$. No ungrounded mechanism or separate residual predictor contributes to the score.

The analysis proceeds in a fixed order. Section~\ref{sec:method} constructs a predictor whose scoring routes are restricted to $\mathcal M$. Sections~\ref{sec:boundary_performance} and~\ref{sec:boundary_validation} verify that it is a sufficiently strong and structurally faithful learnable reference. Section~\ref{sec:covered_performance} attributes the recovered performance to mechanisms and historical facts. Only then do Sections~\ref{sec:neural_advantage} and~\ref{sec:explaining_advantage} locate and explain the remaining neural predictive gain.

\section{LiFTER: Establishing the Known-Mechanism Boundary}
\label{sec:method}

\subsection{Design requirements}
\label{sec:design_conditions}

LiFTER establishes a learnable reference that permits only known temporal interaction mechanisms. It satisfies three requirements. First, every query score must arise from an actual execution of a prespecified mechanism. Second, temporal and candidate compatibility within each mechanism must remain learnable so that a rigid heuristic implementation does not artificially weaken the reference. Third, every score must be decomposable by mechanism and historical fact so that the source of recovered performance can be verified.

The first requirement fixes the comparison boundary, the second estimates it more fairly than fixed heuristics, and the third enables the attribution and interventions in Section~\ref{sec:covered_performance}. The construction therefore serves as the reference for the neural residual analysis in Section~\ref{sec:neural_advantage}.

\begin{center}
\begin{minipage}{\linewidth}
\centering
\includegraphics[width=\linewidth]{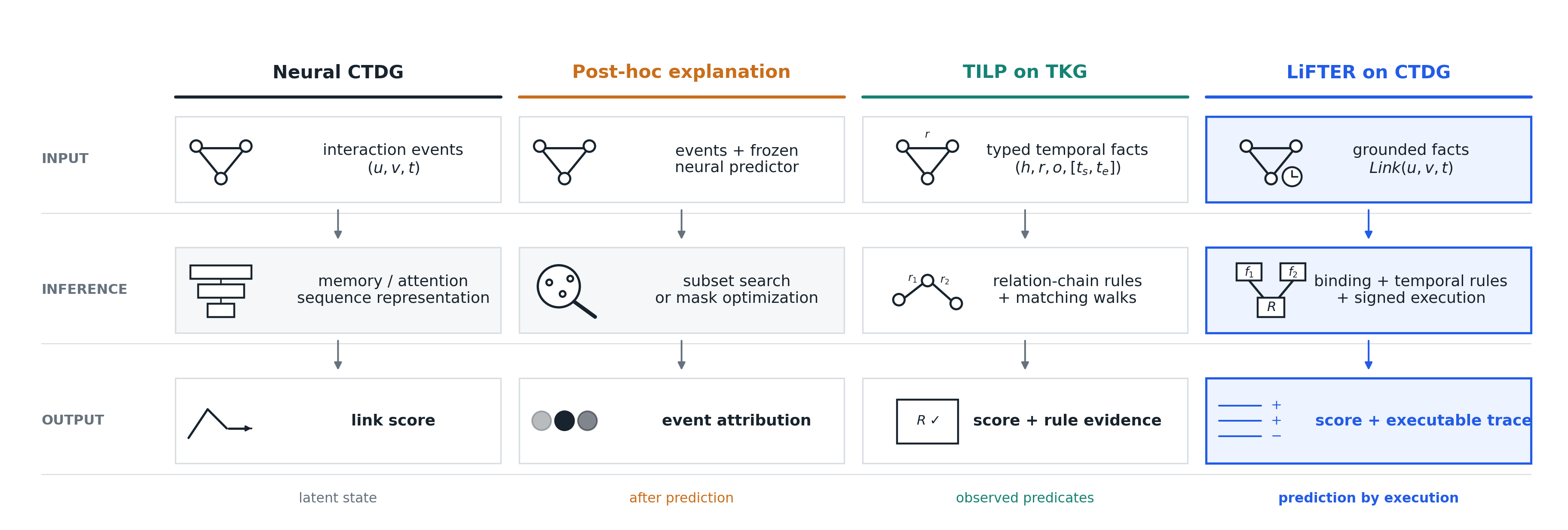}
\captionof{figure}{A conventional neural CTDG model compresses historical events into latent states, whereas a post-hoc explainer searches an existing prediction for important events. Temporal knowledge-graph rule models execute rules that compose observed semantic relations. LiFTER instead represents CTDG interactions without a semantic relation vocabulary as facts preserving entities and timestamps, and computes scores exclusively through prespecified bindings and temporal mechanisms.}
\label{fig:model_comparison}
\end{minipage}
\end{center}

The key distinction in Fig.~\ref{fig:model_comparison} is the computational route rather than the mere availability of an explanation. In LiFTER, executable mechanisms and their participating facts constitute the scoring path itself. We can therefore interpret the performance in Section~\ref{sec:boundary_performance} as a reference achieved under the explicit mechanisms and then compare it with neural models query by query.

\subsection{Fact representation of interactions}
\label{sec:fact_representation}

Each observed interaction is represented as a historical fact $\operatorname{Link}(u,v,t)$. This representation preserves the source entity, destination entity, and timestamp of the original interaction without introducing semantic relations absent from the data.

Entity identifiers first serve symbolic binding conditions. Same-pair recurrence, for example, holds only when both arguments of a historical fact exactly match query $(X,Y)$; embedding similarity does not replace entity identity. Neural parameters evaluate the compatibility and magnitude of an execution only after a valid binding has been established.

For each query, an input index retrieves the 128 most recent events incident to its source or candidate destination. The executor then retains the most recent $H$ facts at each endpoint as grounding candidates. Thus, history length 128 is the incident-event buffer supplied by the index, whereas $H$ limits the number of endpoint-specific facts that may participate in rule grounding. Neither is a temporal threshold, and the original timestamps of selected facts are preserved.

\subsection{Temporal interaction mechanisms}
\label{sec:mechanisms}

The constrained predictor uses the historical context of the query pair and endpoints, same-pair renewal, positioned recurrence, and short sequential transitions. Their execution conditions are collected in Eq.~\eqref{eq:mechanism_bodies}.

\begin{subequations}\label{eq:mechanism_bodies}
\begin{align}
\mathcal B_{\mathrm{unary}}={}&\{P_1(X,Y,T),P_1(Y,X,T),P_1(X,Z,T),
P_1(Z,X,T),P_1(Y,Z,T),P_1(Z,Y,T)\},\quad T<T_q,
\label{eq:unary}\\
\mathcal B_{\mathrm{renew}}={}&P_1(X,Y,T_1)\land P_1(X,Y,T_2),
\quad T_1<T_2<T_q,
\label{eq:renewal}\\
\mathcal B_{\mathrm{pos}}={}&P_1(X,Y,T)\land\operatorname{Pos}_X(T)=j,
\quad 1\le j\le H,
\label{eq:position}\\
\mathcal B_{\mathrm{tr2}}={}&P_1(X,Z_1,T_1)\land P_1(X,Z_2,T_2),
\quad T_1\le T_2<T_q.
\label{eq:transition_body}
\end{align}
\end{subequations}

\textbf{Historical context of the query pair and endpoints.} The six directions in Eq.~\eqref{eq:unary} cover cases in which a historical fact shares an endpoint with the query source or destination. $P_1(X,Y,T)$ is a direct recurrence of the current pair, while other facts incident to $X$ or $Y$ provide local source or candidate context. Equality with a query argument determines the role of each fact, and every binding is checked against concrete entity identity.

\textbf{Same-pair renewal.} When two events of the same pair satisfy Eq.~\eqref{eq:renewal}, the model learns the compatibility between the preceding recurrence interval $T_2-T_1$ and the current query interval $T_q-T_2$. This mechanism uses the temporal structure of recurrence rather than only the existence of a past edge.

\textbf{Positioned recurrence.} Equation~\eqref{eq:position} preserves the position of a past query-pair event in the recent history of source $X$. Here $j=1$ denotes the most recent fact, and every position $1\le j\le H$ has a separate signed weight.

\textbf{Short sequential transitions.} Equation~\eqref{eq:transition_body} preserves the two most recent destinations of source (X) in temporal order; the one-event transition uses only the second fact. Distinct timestamps are temporally ordered, while equal timestamps are treated as simultaneous recent facts. The mechanism executes only when the required events and ordering exist, and destination--candidate compatibility is evaluated only after the executor grounds the body in actual facts.

Let $a$ and $b$ denote $d$-dimensional vectors of recent destinations and the candidate, and let $r_H(X,Y)$ indicate whether the query pair occurs in the recent source history. Under the $K=1$ setting used in the main experiments, transition contributions are
\begin{subequations}\label{eq:transition_potential}
\begin{align}
c_{\mathrm{tr},1}
&=\frac{a_Z^\top b_Y}{\sqrt d}\,s_{1,r_H(X,Y)},
\label{eq:transition_one}\\
c_{\mathrm{tr},2}
&=\frac{s_{2,r_H(X,Y)}}{\sqrt d}\sum_{\ell=1}^{d}
a^{(1)}_{Z_1,\ell}a^{(2)}_{Z_2,\ell}b^{(2)}_{Y,\ell},
\label{eq:transition_two}
\end{align}
\end{subequations}
where $s_{m,r}=\operatorname{softplus}(\rho_m)\operatorname{softplus}(\eta_{m,r})/\log2$ is a learned positive scale. These vectors are accessed only when a grounding of Eq.~\eqref{eq:transition_body} exists, so compatibility alone cannot create an execution.

This construction deliberately retains the mechanisms under study rather than attempting to express arbitrary-length event sequences or global graph structure. Table~\ref{tab:mechanism_specification} summarizes how the rules form the seven components of the final score.
\begin{center}
\begin{minipage}{\linewidth}
\centering
\small
\setlength{\tabcolsep}{3pt}
\begin{tabular}{p{0.17\linewidth}p{0.33\linewidth}p{0.20\linewidth}p{0.22\linewidth}}
\toprule
Component & Execution condition & Input & Contribution \\
\midrule
Direct pair & $P_1(X,Y,T)$ or $P_1(Y,X,T)$, $T<T_q$ & Same-pair fact and time gap & Signed weight on the directional maximum \\
Source context & $P_1(X,Z,T)$ or $P_1(Z,X,T)$ & Source-bound facts and gaps & Signed weight on a directional capped sum \\
Candidate context & $P_1(Y,Z,T)$ or $P_1(Z,Y,T)$ & Candidate-bound facts and gaps & Signed weight on a directional capped sum \\
Pair renewal & Two $P_1(X,Y,T)$ facts with $T_1<T_2<T_q$ & $T_2-T_1$ and $T_q-T_2$ & Capped sum of gap-match evidence over ordered pairs \\
Positioned recurrence & $P_1(X,Y,T)$ and $\operatorname{Pos}_X(T)=j$ & Source-history position $j$ & Sum of learned signed position weights \\
One-event transition & Most recent outgoing $P_1(X,Z,T)$ exists & $Z$, candidate $Y$, recurrence guard & Grounded compatibility in Eq.~\eqref{eq:transition_one} \\
Two-event transition & Two recent outgoing facts with $T_1\le T_2<T_q$ & Ordered $Z_1,Z_2$, candidate $Y$, guard & Grounded compatibility in Eq.~\eqref{eq:transition_two} \\
\bottomrule
\end{tabular}
\captionof{table}{Seven execution components used in the main experiments. Every dataset uses one observed predicate $P_1$ and the same execution conditions.}
\label{tab:mechanism_specification}
\end{minipage}
\end{center}

\subsection{Learnable execution}
\label{sec:learnable_execution}

The symbolic executor determines which historical facts satisfy the bindings and temporal conditions of a mechanism. Neural parameters learn temporal compatibility, candidate compatibility, and the sign of a contribution only for valid executions.

For a valid execution, temporal compatibility between a historical fact $g=(u,v,t)$ and the query is computed with a kernel over log elapsed time:

\begin{equation}
\label{eq:temporal_kernel}
e_m(g,q)=\exp\!\left[-\frac12
\left(\frac{\log(1+T_q-t)-\mu_m}
{\operatorname{softplus}(\lambda_m)+0.05}\right)^2\right].
\end{equation}

In Eq.~\eqref{eq:temporal_kernel}, $\mu_m$ specifies the preferred interval of mechanism $m$, and $\lambda_m$ its tolerated width. Both are learned from data. Same-pair renewal uses $\log(1+T_q-T_2)-\log(1+T_2-T_1)$ in place of absolute recency. The direct-pair unary rule takes a maximum to avoid duplicate counting of the same pair; other unary and renewal rules use the capped sum $E_r=\min(e^4,\sum_g e_r(g,q))$. If no valid grounding exists, $E_r=0$.

An event row satisfying several mechanisms is recorded once for each mechanism, and distinct event rows remain separate facts even when their endpoints and timestamps coincide. Repeated facts within the direct-pair component use the idempotent maximum above. Positioned recurrence sums a position weight over matching facts, while transitions select only the most recent one or two outgoing facts. Signed contributions are summed without additional normalization across mechanism families.

Short sequential transitions use embeddings of actual recent destinations and the candidate to compute compatibility. The result enters the score only under the required historical events and exact source binding; candidate embeddings do not create a separate residual score.

\subsection{Scoring and learning}
\label{sec:score_learning}

Summing contributions from all executions produces a candidate score, which we separate into seven mechanism components:

\begin{equation}
\label{eq:score_components}
s(q)=b+s_{\mathrm{pair}}+s_{\mathrm{src}}+s_{\mathrm{cand}}
+s_{\mathrm{renew}}+s_{\mathrm{pos}}+s_{\mathrm{tr1}}+s_{\mathrm{tr2}}.
\end{equation}

The first three components in Eq.~\eqref{eq:score_components} represent the direct pair, source context, and candidate context. The remaining four represent same-pair renewal, positioned recurrence, one-event transition, and two-event transition. Every nonzero component records the historical facts that executed it, the satisfied binding and temporal conditions, its aggregate evidence, and its signed contribution.

Signed weights of unary and renewal rules combine a term shared by structural schema and predicate slots with a clause-specific term, following the LiFTER specification:
\begin{equation}
w_r=\tanh\!\left[t_{\tau(r)}^\top
\left(\bigodot_{s\in\operatorname{body}(r)}q_{k_s}\right)+\theta_r\right].
\label{eq:clause_weight}
\end{equation}
The main experiments use $K=1$, fixing predicate assignment to $P_1$ for every valid fact; learning in Eq.~\eqref{eq:clause_weight} adjusts schema and clause weights. Grounding and family aggregation use sums without rule-family normalization, except for the direct-pair maximum defined above.

Parameters are trained to raise the score of a true future link and lower that of a sampled negative link using binary cross-entropy:

\begin{equation}
\label{eq:prediction_loss}
\mathcal L_{\mathrm{pred}}
=-\log\sigma(s^+)-\log\sigma(-s^-),
\end{equation}
where $s^+$ and $s^-$ are the scores of the true link and negative candidate. The main experiments treat interactions as a single observation type and use the same mechanism configuration for every dataset. Dataset-specific rules are never selected; learning adjusts temporal compatibility, candidate compatibility, and execution weights.

This realizes the constrained scoring function required by Section~\ref{sec:problem_definition}. Section~\ref{sec:boundary_performance} measures how much neural-baseline performance it recovers, and Section~\ref{sec:boundary_validation} verifies that the recovered performance arises through the execution structure above rather than hidden predictive routes.

\section{Experiments}
\label{sec:experiments}

\subsection{Experimental Setup}
\label{sec:experimental_setup}

We use the first 32,768 events of Wikipedia, Reddit, MOOC, and LastFM, split chronologically into 85\% development and 15\% test intervals. Trainable models are optimized for 10 epochs with seeds 7, 17, and 27. LiFTER uses a predicate vocabulary of $K=1$, an incident-event buffer of 128, hidden dimension 64, and transition dimension 32 on every dataset. The buffer limits the number of past events retrieved by the index at each endpoint, whereas the executor uses only the most recent $H$ as grounding candidates.

We select grounding capacity $H\in\{10,20,40,80\}$ on a chronological validation interval disjoint from the test interval. After reserving the final 15\% for testing, we use the last 15\% of the preceding development prefix for selection. Each candidate $H$ is applied jointly to all four datasets and three seeds. The highest macro-mean historical AUC selects $H=10$, which is then fixed for every test; ties favor the smaller value.

Baselines are EdgeBank, TGN, TGAT, DyGFormer, GraphMixer, CRAFT, and CRAFT-R. Every model uses the same temporal split and positive--negative candidate pairs. We report AUC and AP under both historical and random negatives. Historical negatives are available for 51.1\% of Wikipedia, 48.9\% of Reddit, 97.7\% of MOOC, and 99.7\% of LastFM queries. When no past destination is available, we draw a random candidate from the common destination set. Actual historical candidates and fallbacks are identical across models and seeds.

This common protocol underlies all subsequent comparisons. Section~\ref{sec:boundary_performance} establishes model-level performance, Section~\ref{sec:boundary_validation} verifies its structural source, Section~\ref{sec:covered_performance} decomposes LiFTER scores for matched candidates, and Section~\ref{sec:neural_advantage} directly compares query-level successes of LiFTER and neural models.

\subsection{Establishing the Known-Mechanism Boundary}
\label{sec:boundary_performance}

We first measure performance when prediction is restricted to the known temporal interaction mechanisms. Reporting both negative protocols checks whether the reference is strong only under a particular sampling scheme.

\begin{center}
\begin{minipage}{\linewidth}
\centering
\small
\setlength{\tabcolsep}{5pt}
\begin{tabular}{llrrrr}
\toprule
Dataset & Model & Historical AUC & Historical AP & Random AUC & Random AP \\
\midrule
Wikipedia & LiFTER & 89.20±0.08 & 89.63±0.13 & 97.47±0.07 & 98.04±0.04 \\
 & EdgeBank & 67.08±0.00 & 60.67±0.00 & 92.63±0.06 & 92.46±0.13 \\
 & TGN & \textbf{91.72±0.22} & \textbf{91.78±0.22} & 95.82±0.25 & 96.06±0.24 \\
 & TGAT & 87.33±0.13 & 88.99±0.13 & 97.44±0.13 & 98.01±0.09 \\
 & DyGFormer & 87.17±0.26 & 88.75±0.42 & 97.68±0.18 & 98.17±0.10 \\
 & GraphMixer & 87.40±0.26 & 89.21±0.37 & 97.54±0.21 & 98.09±0.10 \\
 & CRAFT & 85.68±1.17 & 86.04±1.27 & \textbf{97.90±0.09} & \textbf{98.34±0.03} \\
 & CRAFT-R & 76.03±1.49 & 71.48±2.63 & 97.79±0.10 & 98.19±0.11 \\
\midrule
Reddit & \textbf{LiFTER} & \textbf{83.52±0.19} & \textbf{82.72±0.38} & 96.65±0.09 & 97.20±0.07 \\
 & EdgeBank & 63.08±0.00 & 57.92±0.00 & 87.36±0.05 & 87.21±0.10 \\
 & TGN & 81.84±0.05 & 81.10±0.29 & 94.63±0.11 & 94.76±0.20 \\
 & TGAT & 80.38±0.15 & 80.84±0.24 & 96.30±0.16 & 97.08±0.05 \\
 & DyGFormer & 80.46±0.30 & 81.02±0.36 & 96.43±0.28 & 97.16±0.15 \\
 & GraphMixer & 80.49±0.08 & 80.89±0.17 & 96.50±0.12 & 97.21±0.04 \\
 & CRAFT & 79.60±0.57 & 76.95±0.72 & \textbf{97.53±0.07} & \textbf{97.90±0.05} \\
 & CRAFT-R & 68.81±0.66 & 60.73±0.53 & 97.52±0.09 & 97.79±0.04 \\
\midrule
MOOC & LiFTER & 86.72±0.10 & \textbf{85.07±0.08} & 97.05±0.04 & 96.51±0.26 \\
 & EdgeBank & 30.42±0.00 & 42.66±0.00 & 71.86±0.19 & 67.45±0.24 \\
 & TGN & 82.52±0.21 & 77.90±0.13 & 95.97±0.23 & 94.98±0.17 \\
 & TGAT & 83.50±0.51 & 80.08±0.80 & 96.77±0.08 & 96.09±0.09 \\
 & DyGFormer & 85.90±0.76 & 83.37±1.45 & 97.21±0.10 & 96.76±0.18 \\
 & GraphMixer & \textbf{87.09±0.10} & 84.84±0.23 & \textbf{97.45±0.05} & \textbf{96.99±0.06} \\
 & CRAFT & 84.06±0.41 & 79.41±1.15 & 96.87±0.22 & 95.95±0.28 \\
 & CRAFT-R & 82.37±1.75 & 76.43±3.28 & 96.80±0.15 & 95.65±0.49 \\
\midrule
LastFM & \textbf{LiFTER} & \textbf{70.46±0.49} & \textbf{71.75±0.44} & 81.54±0.04 & 84.62±0.08 \\
 & EdgeBank & 38.88±0.00 & 45.14±0.00 & 85.24±0.16 & 82.32±0.27 \\
 & TGN & 61.43±0.87 & 62.98±0.46 & 84.68±0.88 & 85.07±0.85 \\
 & TGAT & 56.31±0.38 & 64.20±0.54 & 90.08±0.18 & 91.82±0.16 \\
 & DyGFormer & 56.85±0.29 & 64.34±0.43 & 90.55±0.20 & 92.12±0.17 \\
 & GraphMixer & 59.59±0.26 & 65.69±0.47 & \textbf{91.41±0.36} & \textbf{92.77±0.26} \\
 & CRAFT & 66.67±0.64 & 65.21±1.39 & 88.09±0.50 & 88.85±0.57 \\
 & CRAFT-R & 55.35±0.70 & 60.08±0.84 & 90.87±0.17 & 91.56±0.08 \\
\bottomrule
\end{tabular}
\captionof{table}{Mean $\pm$ standard deviation over three seeds. Bold indicates the highest value for each dataset and evaluation protocol.}
\label{tab:predictive_performance}
\end{minipage}
\end{center}

Under historical negatives in Table~\ref{tab:predictive_performance}, LiFTER ranks first in five and second in three of the eight dataset--metric combinations. On Wikipedia it trails TGN by 2.52/2.15 AUC/AP points. On MOOC it trails GraphMixer by 0.37/0.23 points. On Reddit and LastFM it exceeds the best neural result. Thus, recurrence, history position, renewal, and short transitions alone approach or exceed strong neural models on the four datasets. This result ensures that the later neural residual is measured against a strong reference.

LiFTER exceeds EdgeBank in AUC by 20.44--56.30 points across datasets. Its ranking therefore uses more than the existence of a past pair: recurrence intervals, positions in source histories, and candidate-conditioned short transitions materially contribute.

With random negatives, Wikipedia, Reddit, and MOOC remain within one point of the best AUC/AP on each dataset. LastFM trails the strongest baseline by 9.87/8.15 points while ranking first under historical negatives. The constrained predictor on LastFM is consequently strongest at discriminating temporal differences among previously observed candidates rather than broadly excluding unrelated destinations.

The known-mechanism reference is therefore competitive with neural models. Section~\ref{sec:boundary_validation} next verifies that this performance arises from the intended structure.

\subsection{Validating the Boundary}
\label{sec:boundary_validation}

Because the remaining analysis relies on LiFTER as a measuring instrument, we test whether its performance depends on the intended grounding and transitions, whether an unconstrained predictor with matched inputs can reproduce it, whether recorded executions exactly reconstruct scores, and whether the instrument is computationally feasible.

We retrain six variants on MOOC and LastFM under the same historical-negative protocol. \texttt{Transition-only} retains one- and two-event transitions; \texttt{clauses without transitions} retains the other grounded clauses. \texttt{Matched MLP} receives the same inputs as LiFTER but learns scores without grounding constraints. \texttt{No exact binding} removes exact entity matching, while \texttt{random binding} randomizes the permitted argument-binding schemas.

\begin{center}
\begin{minipage}{\linewidth}
\centering
\small
\begin{tabular}{llrr}
\toprule
Dataset & Variant & Historical AUC & Historical AP \\
\midrule
MOOC & Full LiFTER & 86.74 & 85.08 \\
 & Transition-only & 87.06 & \textbf{85.25} \\
 & Clauses without transitions & 58.72 & 62.03 \\
 & Matched MLP & \textbf{87.24} & 85.02 \\
 & No exact binding & 86.31 & 83.17 \\
 & Random binding & 86.15 & 82.85 \\
\midrule
LastFM & \textbf{Full LiFTER} & \textbf{70.67} & \textbf{71.87} \\
 & Transition-only & 68.75 & 70.26 \\
 & Clauses without transitions & 58.32 & 62.54 \\
 & Matched MLP & 65.78 & 65.27 \\
 & No exact binding & 69.41 & 70.77 \\
 & Random binding & 69.28 & 70.76 \\
\bottomrule
\end{tabular}
\captionof{table}{All variants are retrained with three seeds. Values are seed-mean AUC and AP under historical negatives.}
\label{tab:boundary_validation}
\end{minipage}
\end{center}

On MOOC, Table~\ref{tab:boundary_validation} shows that transition-only reproduces full LiFTER, whereas removing transitions lowers AP by 23.05 points. Short sequential transitions therefore determine the boundary on this dataset. Removing exact binding lowers AP by 1.91 points and randomizing binding schemas by 2.23 points, confirming that performance also depends on which historical events may execute a transition.

On LastFM, full LiFTER outperforms every constrained variant. It exceeds transition-only by 1.61 AP points and clauses without transitions by 9.33 points, showing complementarity between transitions and other mechanisms. It also exceeds the matched MLP by 6.60 AP points, so the boundary performance cannot be attributed merely to input features and unconstrained neural fitting.

An independent verifier checks that mechanism-level decomposition matches the actual computation. Starting from raw pre-query history and frozen parameters, it recomputes fact existence, entity bindings, temporal order, aggregation, and signed contributions without trusting stored traces. It reconstructs all 19,664 test predictions across the four datasets within tolerance $2\times10^{-5}$, with a maximum score error of $1.31\times10^{-5}$. It detects every injected corruption: missing facts, timestamps moved beyond the query, altered entities, nonexistent executions, and modified contributions.

Thus, later mechanism- and fact-level attributions concern the actual scoring computation. Scalability is also adequate: extending Reddit from 32,768 to 672,447 events maintains indexing throughput near 31,000 events/s and builds the full index in 21.63 s. At $H=10$, execution takes 0.041 ms/query with batch 128 and reaches 101,600 queries/s with batch 2,048. Recomputing after a fact intervention takes 0.138 ms/query.

\subsection{Accounting for the Covered Performance}
\label{sec:covered_performance}

Sections~\ref{sec:boundary_performance} and~\ref{sec:boundary_validation} establish that known mechanisms recover substantial neural performance through the intended execution structure. We now determine what composes that performance. Sections~\ref{sec:mechanism_contributions} and~\ref{sec:query_regimes} analyze mechanism contributions and their variation by query regime; Section~\ref{sec:fact_intervention} tests dependence on grounded historical facts; and Section~\ref{sec:mechanism_budget} measures how many mechanisms are needed to preserve the boundary.

\subsubsection{Mechanism Contributions}
\label{sec:mechanism_contributions}

We evaluate all $2^7=128$ subsets of the seven score components on identical candidate sets. A mechanism's Shapley value averages the AUC or AP change caused by adding it across every ordering. It therefore includes interactions with other mechanisms, and the seven values sum exactly to the difference between the full and empty models.

\begin{center}
\begin{minipage}{\linewidth}
\centering
\includegraphics[width=\linewidth]{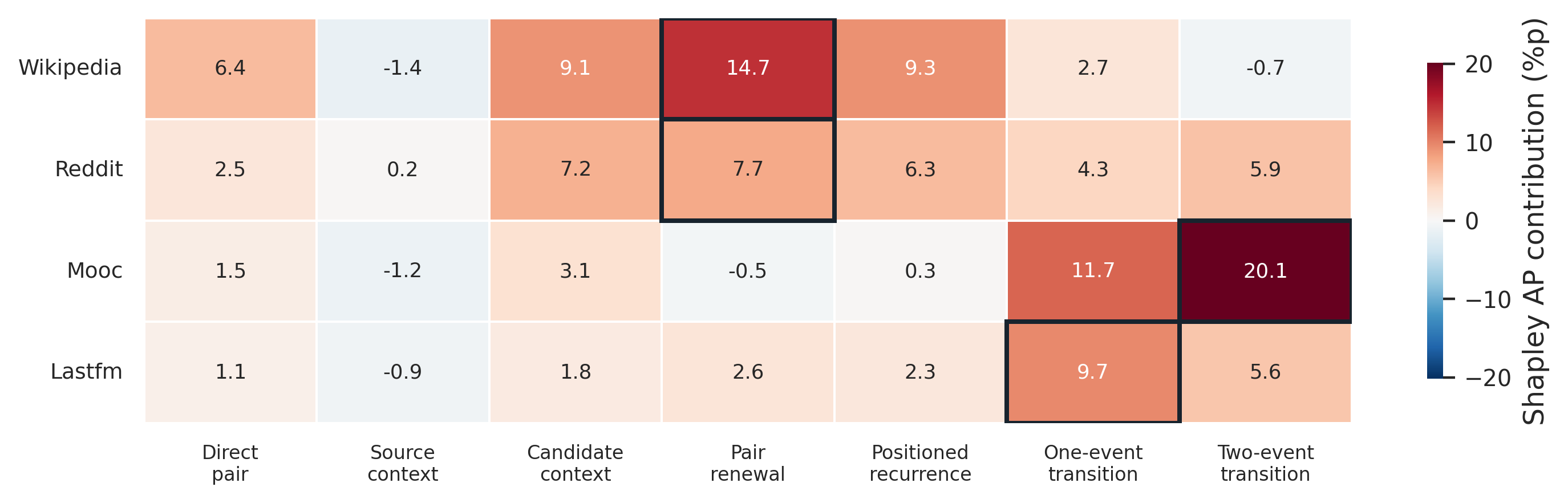}
\captionof{figure}{Historical-AP contribution of each mechanism, computed from all 128 subsets. The dominant source of recovered performance differs by dataset.}
\label{fig:predictive_mechanism_atlas}
\end{minipage}
\end{center}

Figure~\ref{fig:predictive_mechanism_atlas} reveals different dominant mechanisms. Same-pair renewal is most important on Wikipedia and Reddit. The two-event transition from the two most recent destinations dominates MOOC, while the one-event transition from the most recent destination dominates LastFM. Table~\ref{tab:mechanism_deletion} reports the mechanism with the largest AP deletion effect on each dataset.

\begin{center}
\begin{minipage}{\linewidth}
\centering
\begin{tabular}{llr}
\toprule
Dataset & Largest AP deletion effect & AP drop \\
\midrule
Wikipedia & Same-pair renewal & 8.62 points \\
Reddit & Same-pair renewal & 1.23 points \\
MOOC & Two-event transition & 11.76 points \\
LastFM & One-event transition & 4.69 points \\
\bottomrule
\end{tabular}
\captionof{table}{Largest AP deletion effect on each dataset.}
\label{tab:mechanism_deletion}
\end{minipage}
\end{center}

Standalone performance exhibits the same pattern. On MOOC, the two-event transition alone obtains 86.11/85.48 AUC/AP, effectively reproducing the full 86.74/85.08. On LastFM, the one-event transition alone recovers most performance at 68.11/69.65, while the full combination reaches 70.67/71.87. On Wikipedia and Reddit, recurrence, candidate context, and transitions complement one another.

Effects are not additive. On Wikipedia, same-pair renewal and the one-event transition have a positive AP interaction of 4.28 points. On MOOC and LastFM, interactions between one- and two-event transitions are -9.27 and -5.10 points, respectively, indicating competition between temporal resolutions on some queries.

\subsubsection{Variation across Query Regimes}
\label{sec:query_regimes}

Mechanism importance also varies within a dataset. We stratify queries by direct recurrence, source activity, candidate popularity, and early versus late test time, then delete major components. On Wikipedia, renewal effects concentrate on queries with a prior event of the same pair. On Reddit, renewal matters most for highly active sources, while candidate context gains importance for low-activity sources. Removing the two-event transition on MOOC reduces AP by 20.91 points on queries without direct recurrence. The one-event transition remains the most important LastFM component regardless of direct recurrence.

Dataset-level labels are therefore insufficient: the responsible component changes with available history and query conditions. We next intervene on the exact historical facts executing those components.

\subsubsection{Fact-Level Intervention}
\label{sec:fact_intervention}

For each dataset and seed, we select 512 test queries, delete the historical fact with the largest absolute contribution under the dominant mechanism, and rebuild every execution. The control deletes a random historical fact from the same query and evidence range with a matched recency bin.

\begin{center}
\begin{minipage}{\linewidth}
\centering
\small
\begin{tabular}{llrr}
\toprule
Dataset & Target mechanism & Score change top/random & AP drop top/random \\
\midrule
Wikipedia & Same-pair renewal & 1.820 / 1.172 & 0.38 / 0.14 \\
Reddit & Same-pair renewal & 1.949 / 1.571 & -0.06 / 0.52 \\
MOOC & Two-event transition & 1.784 / 0.492 & 13.38 / 1.50 \\
LastFM & One-event transition & 4.490 / 0.484 & 9.61 / 0.42 \\
\bottomrule
\end{tabular}
\captionof{table}{Score changes and AP drops after deleting historical facts supporting the dominant mechanism.}
\label{tab:fact_intervention}
\end{minipage}
\end{center}

In Table~\ref{tab:fact_intervention}, deleting the most influential fact on MOOC and LastFM changes the score 3.63 and 9.29 times more than the control and lowers AP by 13.38 and 9.61 points. Wikipedia also shows larger score and AP changes under targeted deletion. Reddit produces large query-level score changes without a dataset-level AP loss, illustrating that local execution influence and aggregate ranking contribution need not coincide.

Contribution signs agree with interventions: removing a fact that relatively supports the true candidate decreases its margin, whereas removing opposing evidence increases it. Stored contributions therefore encode causal direction in the actual score computation, not merely an importance ranking.

\subsubsection{Mechanism-Budget Performance}
\label{sec:mechanism_budget}

Shapley attribution identifies contributing mechanisms but not how many are required to preserve the boundary. This distinction matters for the residual analysis: rapid saturation would show that a small explicit configuration sustains the reference against which neural advantage is measured.

We split the original held-out interval chronologically while preserving the historical-negative protocol. Its first 50\% is used for subset selection and the second 50\% for final testing. On validation, we evaluate all 128 subsets and select the highest-AP set using at most $k$ mechanisms; the selected set is then fixed on test:

\begin{equation}
\label{eq:mechanism_budget}
\hat S_k=\operatorname*{arg\,max}_{S\subseteq\mathcal M,\,|S|\le k}
v_{\mathrm{val}}(S),
\qquad
T_k=v_{\mathrm{test}}(\hat S_k).
\end{equation}

All subsets use the same frozen checkpoint, queries, and historical negatives. We do not retrain a smaller architecture; unselected mechanism components are simply removed from the decomposed score. The test interval never participates in selection. Accordingly, every point in Fig.~\ref{fig:minimal_mechanism_sets} evaluates the validation-selected $\hat S_k$, and the full-set line uses the same final test interval. Here $k$ is a maximum budget, not necessarily the number selected.

\begin{center}
\begin{minipage}{\linewidth}
\centering
\includegraphics[width=\linewidth]{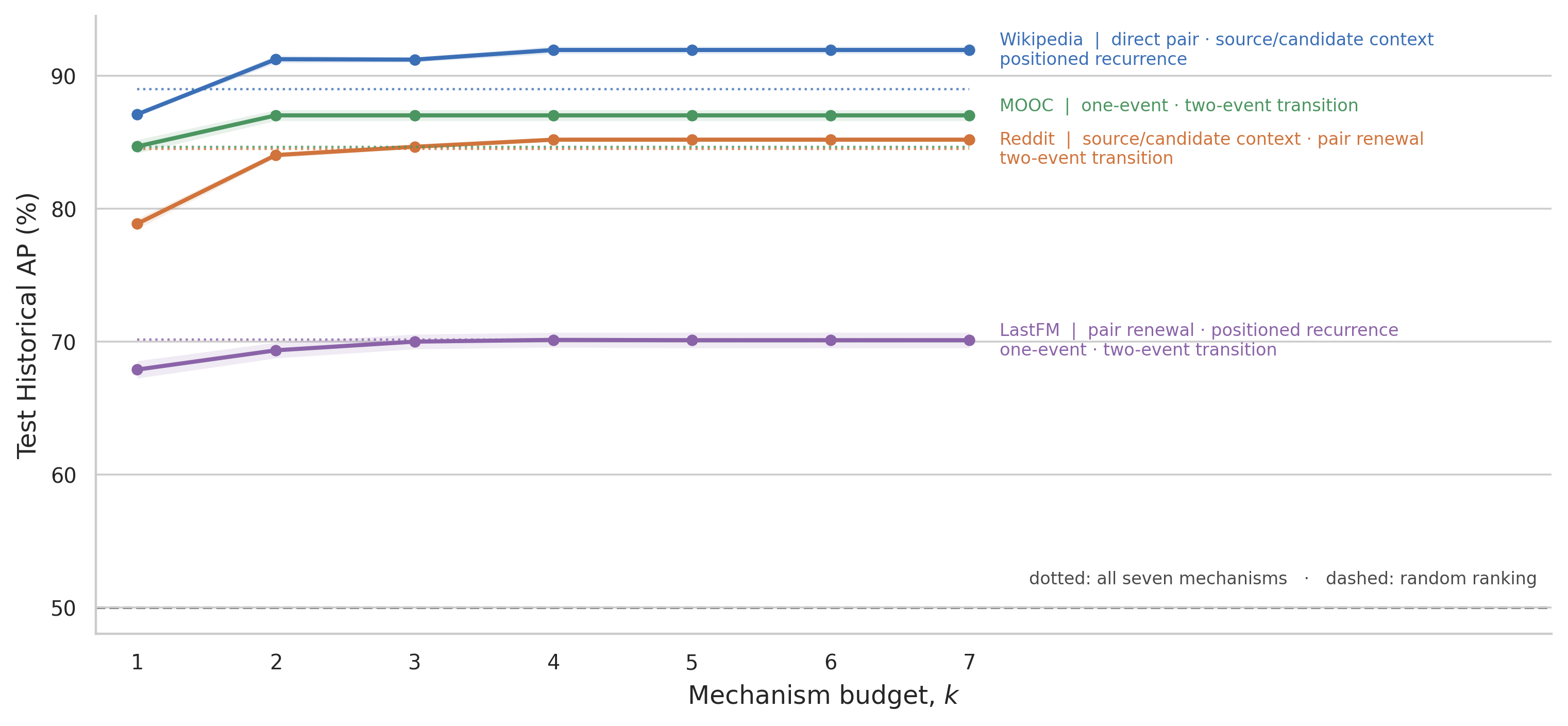}
\captionof{figure}{Test Historical AP of subsets selected on validation under a budget of at most $k$ mechanisms. Curves and bands are the mean and standard deviation over three seeds. Dataset-specific dotted lines show all seven mechanisms on the same test interval; the gray dashed line is random AP (50\%). Right-hand labels show consensus mechanisms selected without a cardinality restriction.}
\label{fig:minimal_mechanism_sets}
\end{minipage}
\end{center}

Performance saturates with two to four mechanisms on every dataset. On Wikipedia, a four-mechanism subset reaches test AP 91.95 versus 88.99 for the full set. Reddit reaches 85.20 versus 84.51. MOOC selects only \texttt{ordered\_transition\_1} and \texttt{ordered\_transition\_2}, reaching 87.03 versus 84.67. On LastFM, a four-mechanism subset reaches 70.13, essentially matching 70.16.

Selected compositions also differ. Wikipedia selects direct pair, source/candidate context, and positioned recurrence. Reddit combines source/candidate context with pair renewal and the two-event transition. MOOC retains only the two sequential transitions. LastFM combines pair renewal and positioned recurrence with both transitions. These compositions recur across most seeds.

The result is specific: LiFTER's recovered performance does not require all seven mechanisms, and the small subset preserving it is dataset-dependent. Neural residuals below nevertheless use full LiFTER to avoid losses induced by subset removal.

\subsection{When Do Neural Models Outperform Known Mechanisms?}
\label{sec:neural_advantage}

\subsubsection{Identifying Neural-Only Success}
\label{sec:neural_only_success}

Aggregate AUC and AP do not reveal whether LiFTER and neural models solve the same queries. We therefore compare their success and failure on each matched test query before locating the neural residual. For model $f$, success means assigning the positive candidate a higher score than the matched historical negative:

\begin{equation}
\label{eq:query_success}
z_f(q)=\mathbb 1\!\left[s_f(q^+)>s_f(q^-)\right].
\end{equation}

For each neural model and full LiFTER, Eq.~\eqref{eq:query_success} partitions queries into both-correct, LiFTER-only, neural-only, and both-wrong sets. We define $z_{\mathrm{neural}}(q)=1$ and $z_{\mathrm{LiFTER}}(q)=0$ as a \textbf{neural-only success}. Separate paired contingency tables are constructed for TGN, GraphMixer, DyGFormer, and CRAFT. Every model receives exactly the same query, positive destination, and historical negative for each seed.

\begin{center}
\begin{minipage}{\linewidth}
\centering
\includegraphics[width=\linewidth]{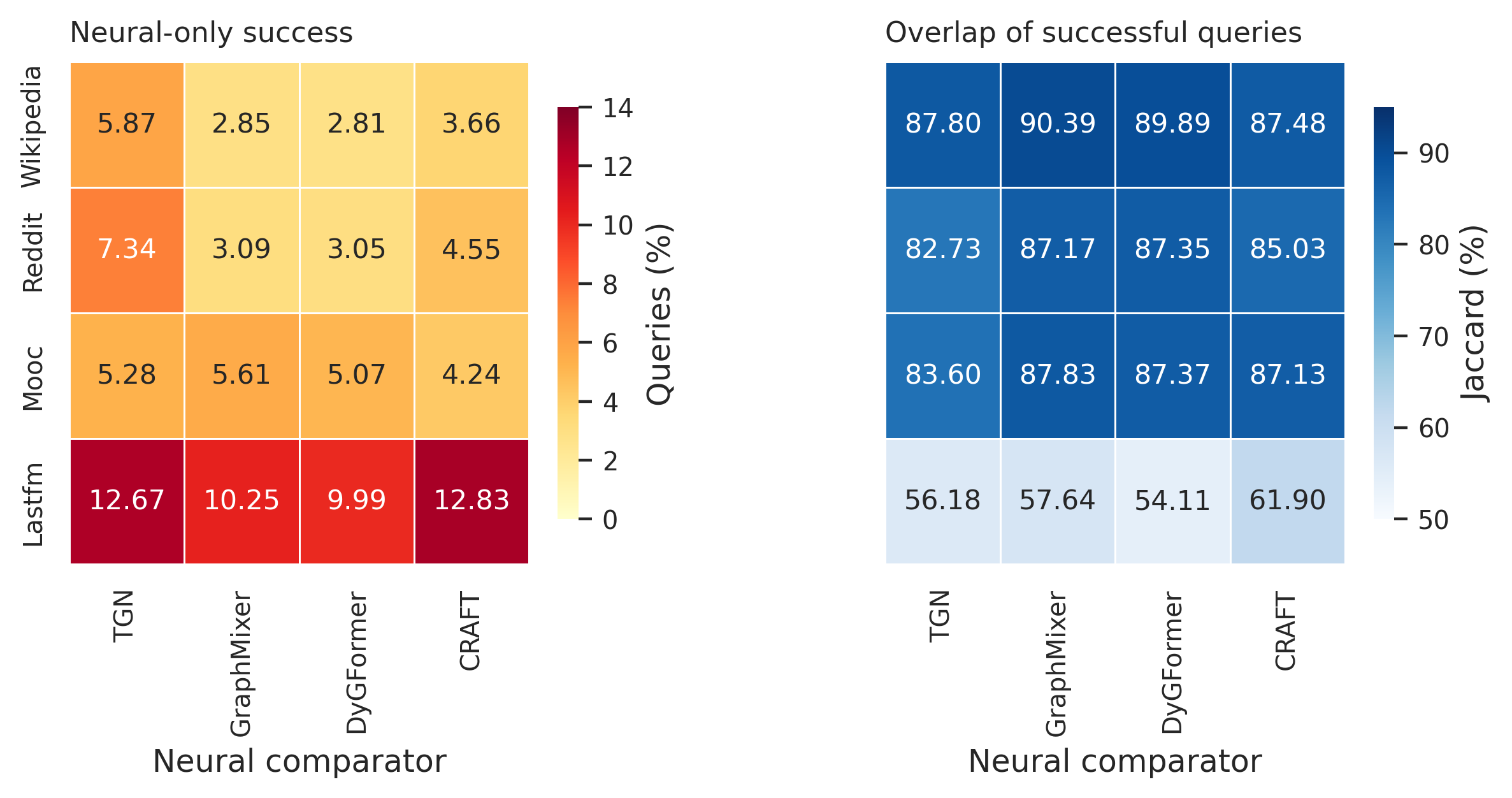}
\captionof{figure}{Query-level comparison between full LiFTER and each neural model. Left: fraction of queries on which LiFTER fails and only the neural model succeeds. Right: Jaccard overlap between their successful-query sets. Values are averaged over seeds 7, 17, and 27.}
\label{fig:neural_residual_overlap}
\end{minipage}
\end{center}

Neural-only successes in Fig.~\ref{fig:neural_residual_overlap} occur for every dataset, neural model, and seed. Their mean rates are 2.81--5.87\% on Wikipedia, 3.05--7.34\% on Reddit, 4.24--5.61\% on MOOC, and 9.99--12.83\% on LastFM. Across 5,000 block-bootstrap samples using blocks of 128 consecutive queries, all 48 comparisons have a 95\% confidence interval above zero; the smallest lower bound is 2.22\%. Jaccard overlap is 82.73--90.39\% on Wikipedia, Reddit, and MOOC, but only 54.11--61.90\% on LastFM.

Near-ties do not explain the result. Removing queries whose normalized absolute margin is at most 0.01 discards up to 4.27\% of queries, yet neural-only success remains at 1.83--12.83\% in every comparison. Thus, a nonempty and reproducible set of neural-only queries exists beyond what aggregate scores expose.

\subsubsection{Locating the Neural Residual}
\label{sec:residual_location}

We test whether these queries arise because few mechanisms are available or because the positive and negative candidates have similar executability profiles. For each test query, $Y^+$ is the true destination and $Y^-$ the evaluation negative. Historical negatives are used when available and random fallbacks otherwise. For each of the seven mechanisms,

\begin{equation}
\label{eq:execution_indicator}
a_m(q,Y)=1
\end{equation}
when mechanism $m$ can execute at least once for candidate $Y$, and zero otherwise. Candidate-profile distance is

\begin{equation}
\label{eq:execution_distance}
D_{\mathrm{exec}}(q)=\sum_{m\in\mathcal M}
\left|a_m(q,Y^+)-a_m(q,Y^-)\right|.
\end{equation}

$D_{\mathrm{exec}}=0$ means that exactly the same mechanism types are executable for both candidates; larger values indicate more candidate-specific mechanisms. We stratify queries into $0,1,2,\ge3$, and also examine asymmetry in direct recurrence and the total number of executable mechanisms. These bins were fixed before inspecting results. $D_{\mathrm{exec}}$ measures only differences in mechanism executability, not final scores, grounding facts, temporal compatibility, or contribution values.

The neural-only rate combines LiFTER's error rate with the rate at which the neural model recovers those errors. Letting $L$ and $N$ denote LiFTER and neural success,

\begin{equation}
\label{eq:residual_decomposition}
P(N=1,L=0\mid D)
=P(L=0\mid D)P(N=1\mid L=0,D).
\end{equation}

We report both factors and their product and separately analyze genuine historical negatives and random fallbacks.

\begin{center}
\begin{minipage}{\linewidth}
\centering
\includegraphics[width=\linewidth]{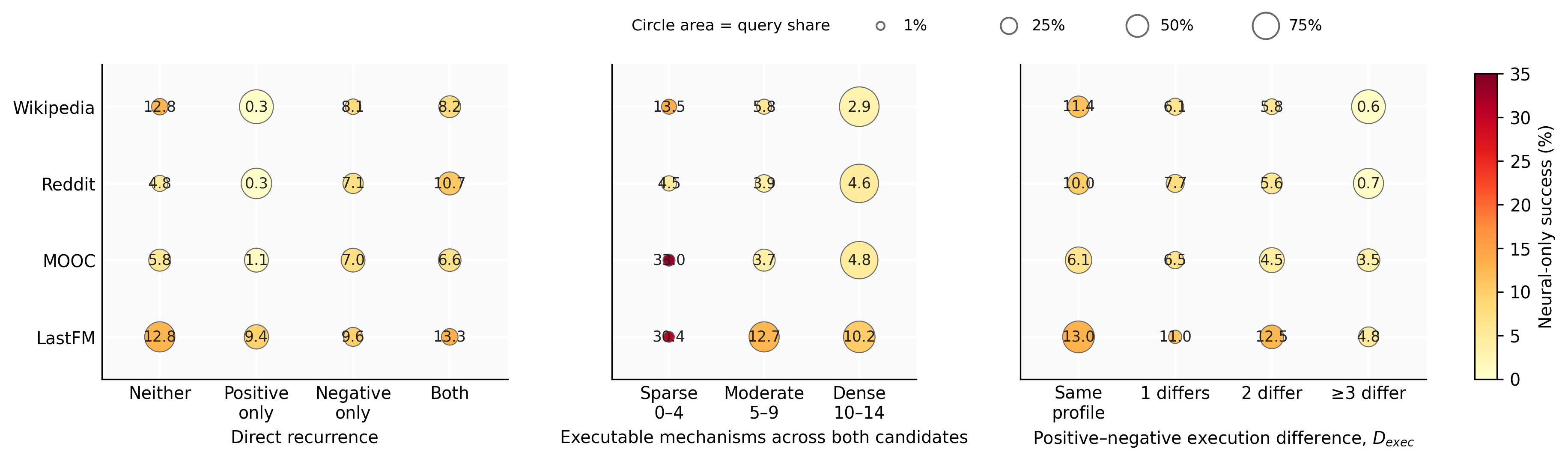}
\captionof{figure}{Numbers and colors show the neural-only success rate averaged over four neural models; circle area shows the fraction of test queries in each regime. Statistical tests retain model- and seed-level pairing and do not treat the displayed model average as independent observations.}
\label{fig:neural_residual_location}
\end{minipage}
\end{center}

Direct recurrence is strongly discriminative in Fig.~\ref{fig:neural_residual_location}. When only the true link has direct recurrence, neural-only success falls to 0.08--0.42\% on Wikipedia and Reddit and 1.02--1.27\% on MOOC. The raw number of executable mechanisms has no consistent relationship with neural-only success.

The most consistent distinction is profile similarity. With $D_{\mathrm{exec}}\le1$, neural-only success is 8.14--12.08\% on Wikipedia, 8.28--10.43\% on Reddit, 5.52--6.72\% on MOOC, and 10.85--15.35\% on LastFM. With $D_{\mathrm{exec}}\ge3$, it drops to 0.19--1.57\%, 0.07--2.20\%, 2.92--3.89\%, and 4.38--5.51\%, respectively. The direction is identical in all 16 dataset--model and 48 seed-level block-bootstrap comparisons.

\begin{center}
\begin{minipage}{\linewidth}
\centering
\small
\setlength{\tabcolsep}{5pt}
\begin{tabular}{lrrr}
\toprule
Dataset & \shortstack{LiFTER error rate\\$D\leq1$ / $D\geq3$} & \shortstack{Conditional recovery\\$D\leq1$ / $D\geq3$} & \shortstack{Neural-only\\$D\leq1$ / $D\geq3$} \\
\midrule
Wikipedia & 23.90 / 16.68 & 35.54 / 10.10 & 8.50 / 1.68 \\
Reddit & 32.23 / 52.54 & 32.40 / 7.09 & 10.44 / 3.73 \\
MOOC & 12.43 / 13.09 & 41.02 / 26.76 & 5.10 / 3.49 \\
LastFM & 34.12 / 13.63 & 37.53 / 35.24 & 12.80 / 4.81 \\
\bottomrule
\end{tabular}
\captionof{table}{Results using genuine historical negatives only, excluding random fallbacks (\%). Conditional recovery and neural-only rates are averaged over four neural models and three seeds. LiFTER's error rate is independent of the neural reference.}
\label{tab:residual_decomposition}
\end{minipage}
\end{center}

Table~\ref{tab:residual_decomposition} shows that the pattern remains after restricting evaluation to genuine historical negatives. Its source differs by dataset. On Wikipedia, both LiFTER error and conditional neural recovery rise. On Reddit and MOOC, conditional recovery rises despite no corresponding increase in low-$D_{\mathrm{exec}}$ LiFTER error. On LastFM, conditional recovery is similar across regimes (37.53\% versus 35.24\%), so most of the neural-only difference comes from LiFTER's error rate.

For fallback queries, Wikipedia and Reddit show no LiFTER errors at $D_{\mathrm{exec}}\ge3$, while fallbacks comprise only 2.0\% and 0.3\% of MOOC and LastFM. The overall pattern is therefore not caused by fallback mixing, although fallback-only regime estimates are too sparse for the same conditional claim.

Neural residuals thus occur more often when the candidates differ little in mechanism executability. This is associated with higher conditional recovery on Wikipedia, Reddit, and MOOC, but primarily with higher LiFTER error on LastFM.

\subsection{What Explains the Neural Advantage?}
\label{sec:explaining_advantage}

We test whether interactions among mechanisms reduce this residual. Let $c_i(q,Y)$ be LiFTER's contribution from mechanism $i$. The linear probe is logistic regression on the seven contributions:

\begin{equation}
\label{eq:linear_probe}
s_{\mathrm{linear}}(q,Y)=\sum_i\beta_i c_i(q,Y).
\end{equation}

It reweights contributions but does not condition one mechanism on another, and therefore serves as the simple-recombination control. The pairwise interaction model additionally crosses a contribution with mechanism executability:

\begin{equation}
\label{eq:pairwise_interaction}
s_{\mathrm{pair}}(q,Y)
=\sum_i\beta_i c_i(q,Y)
+\sum_i\sum_j\gamma_{ij}c_i(q,Y)a_j(q,Y).
\end{equation}

The term $c_i(q,Y)a_j(q,Y)$ remains only when mechanism $j$ is executable, so mechanism $i$'s contribution is interpreted conditionally on co-execution. Both probes use only LiFTER's existing contributions and execution indicators, without new temporal statistics or additional history.

We split the evaluation interval chronologically: the first 25\% fits the probe, the next 25\% selects regularization, the first 50\% then refits it, and the untouched final 50\% provides reported results. Figure~\ref{fig:experimental_timeline} relates this split to model training and $H$ selection. We use the same queries and historical negatives as Section~\ref{sec:neural_advantage}. Confidence intervals come from 5,000 hierarchical bootstrap replicates that first resample seeds and then blocks of 128 consecutive queries.

\begin{center}
\begin{minipage}{\linewidth}
\centering
\includegraphics[width=\linewidth]{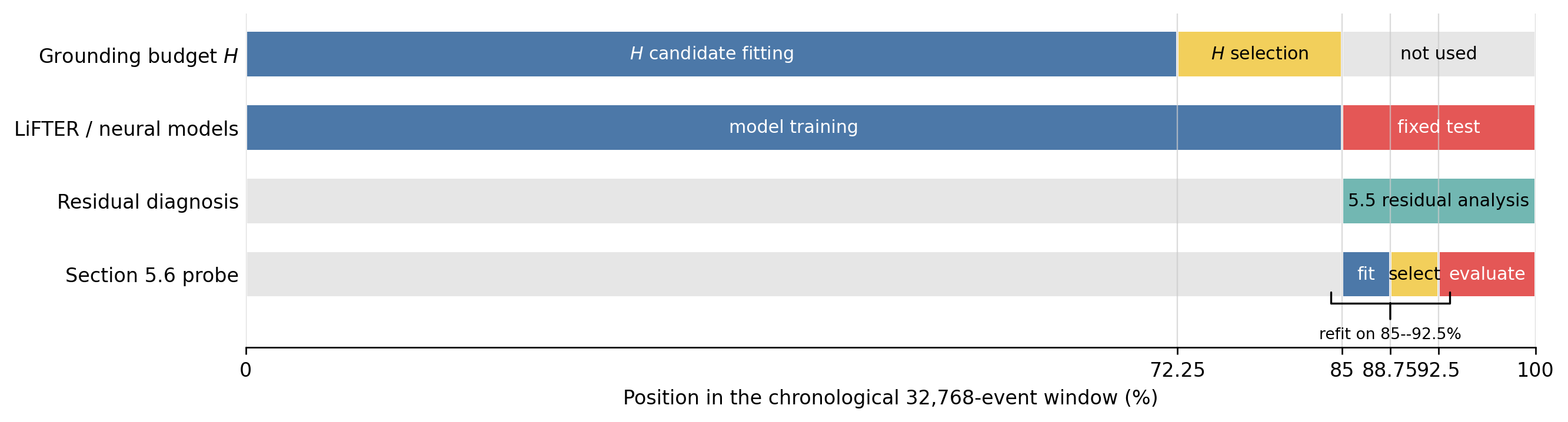}
\captionof{figure}{Temporal allocation of model training, grounding-capacity selection, residual diagnosis, and the Section~\ref{sec:explaining_advantage} probe within the 32,768-event window. $H$ is selected without the final 15\% test interval. The last 50\% of the probe interval is excluded from fitting and regularization selection, but the probe hypothesis was formulated after examining the test traces in Section~\ref{sec:residual_location}. This section is therefore an exploratory post-hoc analysis with temporally separated probe evaluation, not a fully preregistered confirmation.}
\label{fig:experimental_timeline}
\end{minipage}
\end{center}

To distinguish corrected errors from shifted errors, we fix each neural model's original neural-only set relative to LiFTER:
\begin{equation}
R_N=\{q:N(q)=1,\ L(q)=0\}.
\label{eq:original_neural_residual}
\end{equation}
We then count newly recovered queries in $R_N$, newly lost original LiFTER successes, overall paired accuracy, and the residual within $D_{\mathrm{exec}}\le1$.

\begin{center}
\begin{minipage}{\linewidth}
\centering
\small
\setlength{\tabcolsep}{3pt}
\begin{tabular}{llrrrrr}
\toprule
Dataset & Neural reference & LiFTER & Linear probe & Change & \shortstack{Pairwise\\interaction} & Change [95\% CI] \\
\midrule
Wikipedia & TGN & 5.79 & 4.27 & -1.52 & 3.36 & -2.43 [-3.31, -1.64] \\
 & GraphMixer & 2.70 & 3.12 & +0.42 & 2.40 & -0.30 [-0.77, +0.19] \\
 & DyGFormer & 2.69 & 3.00 & +0.31 & 2.37 & -0.31 [-0.80, +0.19] \\
 & CRAFT & 3.68 & 3.34 & -0.34 & 2.78 & -0.90 [-1.39, -0.28] \\
\midrule
Reddit & TGN & 7.35 & 7.74 & +0.39 & 6.14 & -1.21 [-2.04, -0.45] \\
 & GraphMixer & 3.38 & 8.04 & +4.67 & 5.18 & +1.80 [+0.98, +2.52] \\
 & DyGFormer & 3.17 & 7.77 & +4.60 & 4.99 & +1.82 [+1.12, +2.49] \\
 & CRAFT & 4.56 & 7.15 & +2.59 & 5.89 & +1.33 [+0.77, +1.91] \\
\midrule
MOOC & TGN & 5.75 & 7.73 & +1.98 & 4.99 & -0.76 [-1.55, +0.00] \\
 & GraphMixer & 5.86 & 8.03 & +2.17 & 5.02 & -0.84 [-1.54, -0.21] \\
 & DyGFormer & 5.37 & 7.89 & +2.52 & 4.84 & -0.53 [-1.44, +0.26] \\
 & CRAFT & 4.77 & 7.47 & +2.70 & 4.79 & +0.01 [-0.62, +0.64] \\
\midrule
LastFM & TGN & 13.53 & 14.17 & +0.64 & 9.00 & -4.53 [-5.49, -3.60] \\
 & GraphMixer & 11.04 & 11.58 & +0.54 & 6.32 & -4.72 [-5.61, -3.84] \\
 & DyGFormer & 10.54 & 11.03 & +0.49 & 5.86 & -4.68 [-5.60, -3.69] \\
 & CRAFT & 13.75 & 14.59 & +0.84 & 8.81 & -4.94 [-5.81, -4.09] \\
\bottomrule
\end{tabular}
\captionof{table}{Neural-only success rate (\%). Change is the percentage-point difference from original LiFTER; negative values indicate residual reduction. Point estimates are averaged over three seeds, and brackets give 95\% confidence intervals from the chronological block bootstrap.}
\label{tab:neural_residual_interaction}
\end{minipage}
\end{center}

In Table~\ref{tab:neural_residual_interaction}, the linear probe reduces residuals only for TGN and CRAFT on Wikipedia and increases them in most other comparisons. Simple reweighting is therefore inconsistent. The pairwise interaction model reduces all four Wikipedia residuals by 0.30--2.43 points; confidence intervals exclude zero for TGN and CRAFT. On MOOC, point estimates decrease by 0.53--0.84 points for TGN, GraphMixer, and DyGFormer, with only GraphMixer excluding zero. CRAFT is unchanged.

The largest and most consistent effect occurs on LastFM: residuals against all four neural models fall by 4.53--4.94 points, with every interval below zero. Reddit is mixed: TGN decreases by 1.21 points, whereas GraphMixer, DyGFormer, and CRAFT increase by 1.33--1.82 points, all with intervals excluding zero. Conditional interaction is therefore effective throughout LastFM and in parts of Wikipedia and MOOC, but is not a universal explanation.

\begin{center}
\begin{minipage}{\linewidth}
\centering
\small
\setlength{\tabcolsep}{3pt}
\begin{tabular}{llrrrr}
\toprule
Dataset & Neural reference & \shortstack{Recovered\\$R_N$} & \shortstack{Lost LiFTER\\successes} & \shortstack{Overall paired\\accuracy} & \shortstack{$D_{\mathrm{exec}}\le1$\\residual} \\
\midrule
Wikipedia & TGN & 100.3 (70.5\%) & 60.3 (2.9\%) & 85.39 $\rightarrow$ 91.00 & 10.82 $\rightarrow$ 6.13 \\
 & GraphMixer & 48.3 (73.1\%) & 60.3 (2.9\%) & 85.39 $\rightarrow$ 91.00 & 7.14 $\rightarrow$ 4.71 \\
 & DyGFormer & 48.0 (73.2\%) & 60.3 (2.9\%) & 85.39 $\rightarrow$ 91.00 & 7.30 $\rightarrow$ 4.75 \\
 & CRAFT & 61.7 (68.1\%) & 60.3 (2.9\%) & 85.39 $\rightarrow$ 91.00 & 9.35 $\rightarrow$ 5.67 \\
\midrule
Reddit & TGN & 121.0 (67.1\%) & 121.0 (6.4\%) & 76.86 $\rightarrow$ 85.64 & 9.80 $\rightarrow$ 11.24 \\
 & GraphMixer & 46.0 (54.9\%) & 121.0 (6.4\%) & 76.86 $\rightarrow$ 85.64 & 8.99 $\rightarrow$ 7.72 \\
 & DyGFormer & 43.7 (55.7\%) & 121.0 (6.4\%) & 76.86 $\rightarrow$ 85.64 & 8.41 $\rightarrow$ 7.23 \\
 & CRAFT & 61.7 (55.2\%) & 121.0 (6.4\%) & 76.86 $\rightarrow$ 85.64 & 10.50 $\rightarrow$ 10.46 \\
\midrule
MOOC & TGN & 94.7 (67.0\%) & 91.0 (4.3\%) & 85.56 $\rightarrow$ 91.70 & 7.41 $\rightarrow$ 6.75 \\
 & GraphMixer & 100.0 (69.5\%) & 91.0 (4.3\%) & 85.56 $\rightarrow$ 91.70 & 7.50 $\rightarrow$ 7.14 \\
 & DyGFormer & 89.7 (67.7\%) & 91.0 (4.3\%) & 85.56 $\rightarrow$ 91.70 & 7.12 $\rightarrow$ 6.93 \\
 & CRAFT & 77.3 (65.7\%) & 91.0 (4.3\%) & 85.56 $\rightarrow$ 91.70 & 6.34 $\rightarrow$ 6.55 \\
\midrule
LastFM & TGN & 205.0 (61.6\%) & 194.3 (11.6\%) & 68.16 $\rightarrow$ 79.40 & 14.65 $\rightarrow$ 11.31 \\
 & GraphMixer & 181.7 (67.0\%) & 194.3 (11.6\%) & 68.16 $\rightarrow$ 79.40 & 11.68 $\rightarrow$ 8.69 \\
 & DyGFormer & 175.0 (67.5\%) & 194.3 (11.6\%) & 68.16 $\rightarrow$ 79.40 & 11.03 $\rightarrow$ 7.78 \\
 & CRAFT & 217.3 (64.3\%) & 194.3 (11.6\%) & 68.16 $\rightarrow$ 79.40 & 16.07 $\rightarrow$ 12.01 \\
\bottomrule
\end{tabular}
\captionof{table}{Prediction changes induced by the pairwise interaction model. Recovered $R_N$ gives the number and fraction of original neural-only queries newly solved; lost LiFTER successes gives the number and fraction of original LiFTER successes newly missed. Counts are means over three seeds. The final columns show changes from LiFTER to the interaction model in overall paired accuracy and the neural-only rate (\%) for $D_{\mathrm{exec}}\le1$.}
\label{tab:interaction_error_movement}
\end{minipage}
\end{center}

Table~\ref{tab:interaction_error_movement} shows that the interaction model recovers 54.9--73.2\% of the original $R_N$ while losing 2.9--11.6\% of original LiFTER successes. Accounting for both directions, overall paired accuracy rises from 85.39\% to 91.00\% on Wikipedia, 76.86\% to 85.64\% on Reddit, 85.56\% to 91.70\% on MOOC, and 68.16\% to 79.40\% on LastFM. Conditional interactions therefore recover a substantial portion of the neural residual with a net improvement rather than merely exchanging successful queries.

Within $D_{\mathrm{exec}}\le1$, residuals decrease against every neural baseline on Wikipedia and LastFM. MOOC shows smaller reductions and a 0.21-point increase for CRAFT. Reddit decreases for GraphMixer and DyGFormer, changes little for CRAFT, and increases by 1.44 points for TGN. The location identified in Section~\ref{sec:residual_location} therefore aligns most clearly with the interaction effect on Wikipedia and LastFM.

\section{Discussion and Limitations}
\label{sec:discussion}

The central result is that the additional predictive power of neural CTDG models need not remain an opaque aggregate gap. Explicit mechanisms---recurrence, history position, renewal, and short transitions---already account for a substantial portion of strong neural performance, and this coverage quickly saturates with a small, dataset-specific subset. Yet matched candidate comparisons consistently leave neural-only successes. The neural residual is therefore a concrete object of analysis separating known-mechanism coverage from additional predictive gain.

Its location is also structured. Neural-only successes concentrate where the positive and negative candidates have similar mechanism-execution profiles, rather than simply where few mechanisms execute. Additional neural power appears more often where existing mechanisms fail to discriminate competing candidates than where explicit evidence disappears entirely. Neural complexity may consequently matter not only by adding temporal patterns, but also by interpreting and combining several concurrent signals.

Section~\ref{sec:explaining_advantage} provides functional evidence for this interpretation. Simple reweighting does not consistently reduce the residual, whereas conditioning one mechanism's contribution on another's execution recovers many original neural-only queries. Although this change loses some original LiFTER successes, overall paired accuracy increases on every dataset. The interaction probe thus reduces prediction error through more flexible coordination of available evidence rather than merely relocating successes.

From this perspective, neural complexity can act as connective capacity among recurrence, recency, transitions, and candidate context. The meaning of identical transition evidence may change with recurrence or candidate context. The probe does not reconstruct the internal computation of a neural model, but demonstrates that conditional coordination alone can functionally reproduce part of its advantage.

This view also offers a common lens on TGN, GraphMixer, DyGFormer, and CRAFT. Despite distinct memory updates, nonlinear mixing, attention, and candidate-conditioned comparisons, each obtains additional successes on queries that the explicit reference cannot clearly distinguish. Rather than attributing every gain to a particular architectural module, part of the shared benefit may lie in context-dependent coordination of existing temporal evidence.

The implication for model design is direct. Capacity may be allocated selectively to regimes where several mechanisms overlap or conflict, instead of uniformly increasing history length, motif inventories, or latent-state size. Future architectures could explicitly distinguish when strong low-dimensional mechanisms are sufficient and when they require conditional coordination.

The pattern is not identical across datasets. The association between low execution separation and interaction effects is most consistent on Wikipedia and LastFM, partial on MOOC, and model-dependent on Reddit. Moreover, concentration of neural-only success reflects different mixtures of LiFTER error and conditional neural recovery. Neural residuals should therefore be understood as a family of remaining gains produced by several ways in which the explicit reference becomes insufficient.

The known-mechanism boundary is not an upper bound over all implementations of these mechanisms or a complete basis for CTDG dynamics. It is a learnable reference defined by LiFTER's prespecified mechanisms, input range, execution procedure, and scoring design. Likewise, successful interaction probes do not establish that neural models internally implement the same computation. Our contribution is a diagnostic separation of covered performance and neural residuals, together with a testable functional explanation for part of that residual.

Finally, the analysis covers four CTDG datasets and primarily historical-negative link prediction. Longer event streams, inductive settings, other negative schemes, and other domains may alter both coverage and residual location. Applying the same decomposition more broadly may reveal whether residuals beyond conditional interaction arise from long-term memory, higher-order structure, or other forms of contextual dependence.

\section{Conclusion}
\label{sec:conclusion}

We decomposed the additional predictive power of neural CTDG models into performance covered by known temporal interaction mechanisms and the remaining neural residual. Recurrence, history position, renewal, and short sequential transitions recovered a substantial portion of strong neural performance, and this coverage quickly saturated with a small, dataset-specific set of explicit mechanisms.

At the query level, neural-only successes concentrated where positive and negative candidates had similar mechanism-execution profiles. Conditional interactions that adjusted one mechanism's contribution according to another's execution recovered a substantial fraction of these queries and improved overall paired accuracy more effectively than simple reweighting.

Part of neural advantage in CTDG prediction can therefore arise from context-dependent coordination of known, low-dimensional mechanisms as well as from additional temporal patterns. By locating this advantage at the query level and providing a testable functional explanation, our analysis suggests that future neural capacity should distinguish both when existing mechanisms suffice and when they require coordination.

\section*{Acknowledgements}
The authors have no acknowledgements to declare.

\section*{Funding}
This research received no specific grant from funding agencies in the public,
commercial, or not-for-profit sectors.

\section*{Conflict of interest}
The authors declare no conflicts of interest.

\section*{Data availability}
This study uses the public Wikipedia, Reddit, MOOC, and LastFM temporal
interaction datasets distributed with JODIE. The raw data are available from
the Stanford SNAP JODIE repository. The accompanying materialization pipeline
records the source metadata and produces the chronological splits used in this
study.

\section*{Code availability}
The implementation, experiment configurations, result summaries, and scripts
for reproducing the tables and figures are publicly available at
\url{https://github.com/SnowyPainter/LiFTER-public}.

\bibliographystyle{cas-model2-names}
\bibliography{references}

\end{document}